# VPH+ and MPC Combined Collision Avoidance for Unmanned Ground Vehicle in Unknown Environment

Kai Liu, Jianwei Gong, Huiyan Chen

*Abstract*— there are many situations for which an unmanned ground vehicle has to work with only partial observability of the environment. Therefore, a feasible nonholonomic obstacle avoidance and target tracking action must be generated immediately based on real-time perceptual information. This paper presents a robust approach by integrating VPH+ (enchanced vector polar histogram) and MPC (model predictive control). VPH+ is aploited to calculate the desired direction for its environment perception ability and computational efficiency, while MPC is explored to perform a constrained model-predictive trajectory generation. This approach can be implemented in a reactive controller. Simulation experiments are performed in VREP to validate the proposed approach.

I. INTRODUCTION

Unmanned ground vehicle usually has to work in unknown environment with only partial observability, so the capabilities of localization, obstacle avoidance, and reaching a target are essentially needed. Such functions ensure that a UGV platform navigates safely around the obstacles while trying to reach its goal. These fundamental functions are not independently operated but intimately connected to one another. Considering this perspective, many reactive collision avoidance methods have been introduced based on real-time perceptual information.

*A. Piror work*

Some representative examples of reactive approachs are Potential Field Method (PFM), Vector Field Histogram (VFH), Vector Polar histogram (VPH) and enchanced Vector Polar histogram (VPH+).

PFM, inducing attractive force for a target and repulsive force for obstacles, is one of the classical methods and still exploited due to its comfortable implementation. It has, however, some limitations such as getting stuck in a local minima and unable to grasp the environmental connectivity [1].

Based on the concep of PFM and polar histogram using certainty grid map obtained from sensor measurements, VFH and VFH+ are computationally efficient and suitable for real-time operation. However, they still have some shortcomings such as lack of adaptability for complex environments and heavily experience reliance in parameters selection [2, 3].

As the combination of PFM and VFH+ method, the VPH and its enchanced version VPH+ have been proposed to be specialized for high accuracy sensors [4, 5]. VPH+ can obviously reduce the computation burden and make the UGV move more safely at a high speed by taking UGV acceleration and speed into account in the threshold function.

These methods require only partial observability of the environment at the cost of guaranteeing only local optimality. In order to address the local minima problem, a robust VPH+ algorithm is proposed [6-8]. The proposed method is much more suitable for online planning of obstacle avoidance for its advantages of less computations and fast response.

The above cited approaches can be often easily implemented in the form of reactive controllers. However, one of the major drawback of theese methods is that they don't take the multi-constraints and the handling limits of the UGV into consideration. Therefore, these methods are more suitable for mobile robots rather than UGV.

There are also some algorithms generate UGV control actions with vehicle constraints and limits in consideration. Jiangyan proposed a differential constraints-based algorithm for UGV's path tracking and obstacle avoidance [9, 10]. This approach generates a bounch of quintic curves connecting vehicle state with the goal configuration. Then select the obstacle free path as the desired path. The approach assures the continutity of trajectory curvature. However, it usually requires a safe pre-defined reference which is not always available.

Model Predictive Control (MPC) has also received on-going interest in trajectory tracking and obstacle avoidance for its advantages of systematically handling system nonlinearities and multi-constraints, working in a wide operating region and close to the set of admissible states and inputs. Falcone, et al. applied the MPC-based trajectory generating and obstacle aviodance methods in autonomous vehicle control [11-14]. However, it has following shortages when applied in unknown environment.

➢ This approach might be limited in application for the computation time required. The complexity of the planning problem is further increased when obstacles are considered.

➢ They use a minimized cycle to represent the obstacles which is hard to know in real application and it is not suitable for long large obstacles such as walls and so on.

*B. Approach*

For the disadvantages and shortcomings of the past research method, this paper presents an obstacle avoiding and

*Resrach supported by NSFC (51275041).

Kai Liu is with the Intelligent Vehicle Research Center, School of Mechanical Engineering, Beijing Institute of Technology, Beijing, CO 100081 China (e-mail:leoking1025@gmail.com)

Jianwei Gong, Yan Jiang, Yinjian Sun are with the Intelligent Vehicle Research Center, School of Mechanical Engineering, Beijing Institute of Technology, Beijing, CO 100081 China (e-mail:gongjianwei@bit.edu.cn), jiangyan@bit.edu.cn).

goal tracking system based on VPH+ and MPC. VPH+ is adopted to realize obstacle detection and environment modeling in real-time. Meanwhile, it can also generate a desired driving direction which navigates the UGV to the goal while avoiding collision. As the desired direction does not always accord with the multi-constraints and nonlinearities of UGV, MPC is explored to generate a sequence of feasible actions that leading the UGV to the desired direction. By integrating VPH+ algorithm with MPC, the UGV could feasibly avoid obstacles and aiming towards the target after obstacle avoidance during path following.

*C. Layout*

The paper is organized into five sections. In Section 2 we introduce the scheme of MPC and develop the kinetimatic vehicle model used to predict and optimize the future system behavior. In Section 3 we introduce the VPH+ and MPC based controller and develop the computational method of solution. In Section 4 we present the simulation results that are carried out in V-Rep environment and in Section 5 the conclusion.

## II. PROBLEM STATEMENT

The calculated results of VPH+ are not always meet the constraint of vehicle. In this paper we address the problem using MPC, achieving the goal direction while respecting the dynamic and nonholonomic limitation on UGV.

The idea of MPC is to utilize a UGV model in order to predict and optimize its future behavior. It is an optimazation based method for the feedback control of UGV. We predict the UGV's future trajectory in $N_p$ (predict horizon) steps ahead of time. For each follower in the formation, the current control action is obtained by solving online, at each sampling instant, the distributed cost function using the current state of the UGV as the initial state. The optimization yields an optimal control sequence and the first $N_c$ (control horizon) elements of this sequence are applied to the system until the next sampling instant.

As this research focuses mainly on the impact of changing the steering on the motion of the UGV. Therefore, kinematic modeling of the UGV is more suited as compared to dynamic modeling. A UGV model made up of a rigid body and non deforming wheels is considered, as shown in Fig. 1. It is assumed that the UGV moves on a plane without slipping, i.e., there is a pure rolling contact between the wheels and the ground.

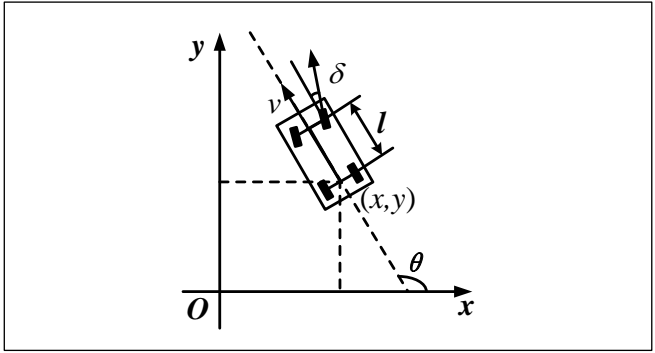

Figure 1. Vehicle kinematic model

Under these assumptions, the nonlinear rear-wheeled kinematic model of the UGV is given as follows:

$$\begin{bmatrix} \dot{x} \\ \dot{y} \\ \dot{\theta} \end{bmatrix} = \begin{bmatrix} \cos\theta \\ \sin\theta \\ \dfrac{\tan\delta}{l} \end{bmatrix} v \qquad (1)$$

Or, in a more compact form as:

$$\dot{x} = f(x,u) \qquad (2)$$

Where the state $\chi$ describes the position and orientation of the center of UGV with respect to the Cartesian coordinates of global inertial frame. And the control input $u$ is denoted as the front wheel angle.

For linearization, a linear error model is obtained using Taylor series with respect to a reference trajectory. To do so, consider a reference trajectory also described by (2). Its trajectory $x_r$ and $u_r$ are related by:

$$\dot{x}_r = f(x_r, u_r) \qquad (3)$$

Hence, the linear error model is described as (3).

$$\begin{bmatrix} \dot{x} - \dot{x}_r \\ \dot{y} - \dot{y}_r \\ \dot{\theta} - \dot{\theta}_r \end{bmatrix} = \begin{bmatrix} 0 & 0 & -v_r * \sin\theta_r \\ 0 & 0 & v_r * \cos\theta_r \\ 0 & 0 & 0 \end{bmatrix} \begin{bmatrix} x - x_r \\ y - y_r \\ \theta - \theta_r \end{bmatrix}$$
$$+ \begin{bmatrix} 0 \\ 0 \\ \dfrac{v_r}{L\cos(\delta_r)} \end{bmatrix} [\delta - \delta_r] \qquad (4)$$

In general, since the MPC algorithm is calculated in discrete time, it is necessary to transform the differentail kinematic model into discrete format. Utilizing the Feed-forward difference method (Euler's approximation) to (4), the discretized kinematic model can be obtained as shown in (5) ~ (7).

$$\tilde{\chi}(k+1) = A_{k,t}\tilde{\chi}(k) + B_{k,t}\tilde{u}(k) \qquad (5)$$

$$A_{k,t} = \begin{bmatrix} 1 & 0 & -v_r \sin\theta_r * T \\ 0 & 1 & v_r \cos\theta_r * T \\ 0 & 0 & 1 \end{bmatrix} \qquad (6)$$

$$B_{k,t} = \begin{bmatrix} 0 \\ 0 \\ \dfrac{v_r * T}{L * \cos^2(\delta_r)} \end{bmatrix} \qquad (7)$$

Where, *T* is the sampling time, and *k* is the sampling instant, $v_r$ is the given velocity, and *L* is the wheelbase. In this paper, the reference set point is set to be the origin.

### III. CONTROLLER DESIGN

First, a robust VPH+ is adopted to find the desired direction. In every planning cycle, VPH+ includes five steps: modification of original information, clustering and envrionment modeling, construction of symbol function, construction of threshold function and construction of cost function. Then a MPC trajectory generate and tracking controller is developed.

#### A. Modification of Original Information

The original obstacle information measured by laser range finder (LRF) should be modified to the actual distance that the UGV could reach in each direction. Suppose that the LRF could scan every other degree from 0 to 180, as shown in Fig. 2. $O_i$ is the obstacle point detected by the LRF in the direction *i* (*i*=0, 1, 2, …, 180), and the distance in that direction is $d_i$. $P_r$ is the position of the UGV and the minimum radius of UGV is *R*.

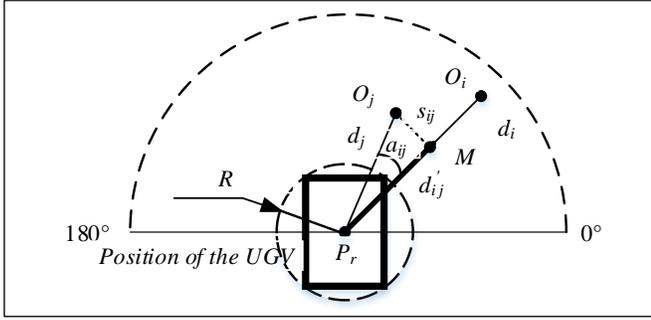

Figure 2. Laser range finder data modification

Assume $O_j$ is any other obstacle point detected by LRF, and its projection point on the line $P_rO_i$ is *M*. the length of line $O_jM$ is $s_{ij}$. Due to the existence of $O_j$, the distance of $O_i$ to the UGV is modified to $d_{ij}'$ (the length of line *PrM*), as shown in (8)~(9). Consider the minimum radius of the UGV, the farthest distance $D_i$ that the UGV could reach in direction *i* is defined by (10).

$$d_{ij}' = \begin{cases} d_i, & s_{ij} > R \\ d_i, (s_{ij} \le R) \text{ and } (d_j \cos(\alpha_{ij}) > d_i) \\ d_j \cos(\alpha_{ij}), \text{others} \end{cases} \quad (8)$$

$$s_{ij} = d_j \sin(y_{ij}) \quad (9)$$

$$D_i = \min(d_{ij}') - R \quad (10)$$

#### B. Obstacle Clustering and enironment modeling

Clustering obstacle points into blocks is an easy way to simplify the environment modeling in real-time. A dynamic window is adopted to reduce the computational burden. The principle of clustering is as follows:

Suppose that *L* is the radius of the rolling dynamic window and $D_i$ is the distance of the $i^{th}$ detection. When $D_i < L$, rolling window will be considered that there are obstacles existing inside in the process of detecting obstacle.

And $d_{i,i+1}$ is the distance between the two adjoining detected points, *γ* is the angular resolution of the LRF. *R_Thr* is the threshold distance that decides whether the two obstacles belong to the same block or not, *R_Thr*=*R*+2Δ*d*. Δ*d* is the puffed distance of UGV. The bigger Δ*d* is, the higher the safety coefficient of obstacle avoidance is. With the comparison of $d_{i,i+1}$ and *R_Thr*, it is determined whether the obstacle belongs to the same block or not.

$$d_{i,i+1} = \sqrt{D_i + D_{i+1} - 2D_iD_{i+1}\cos(\gamma)} \quad (11)$$

After clustering the obstacle points into blocks, a merge of blocks is carried out. A typical block merge situation is shown in Fig. 3. Δ$d_1$, Δ$d_2$ and Δ$d_3$ are the distances of two clustered block ends.

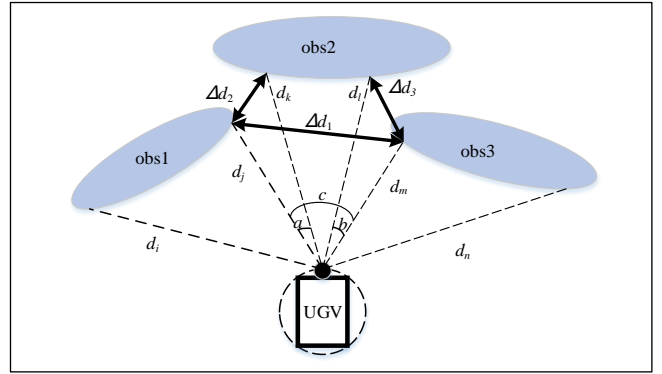

Figure 3. Schematic diagram of clustering

If Δ$d_1$<*R_Thr*, the UGV can't get through between obs1 and obs3, then obs1 and obs3 are seen as the same obstacle and do the merge processing.

If *R_Thr* < Δ$d_1$, the UGV can get through between obs1 and obs3, then it is the time to judge whether there are feasibility paths between obs1 and obs2, obs2 and obs3.

If Δ$d_2$<*R_Thr* and Δ$d_3$< *R_Thr* , there are no feasibility paths between obs1 and obs2 , obs2 and obs3 ,then obs1 , obs2 and obs3 are seen as an obstacle and do the merge processing.

If Δ$d_2$<*R_Thr* and Δ$d_3$>*R_Thr*, the UGV will get through between obs2 and obs3. If Δ$d_3$< *R_Thr* and Δ$d_2$> *R_Thr*, the UGV will get through between obs1 and obs2.

If Δ$d_2$>*R_Thr* and Δ$d_3$>*R_Thr*, obs1, obs2 and obs3 are independent and the UGV can get through between them freely.

#### C. Construction of Symbol Function

Symbol function is constructed by comparing each clustered block with its adjacent blocks. It denotes what kind of block an obstacle point belongs to after LRF data are clustered. If the distance of both ends of the block to the UGV are shorter than that of their adjacent block's ends, then this block is concave, vice verse.

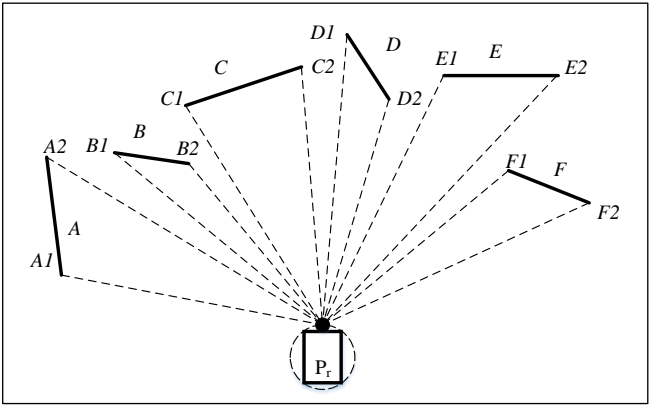

Figure 4. Obstacle block classification

For example, as shown in Fig. 4, there are 6 clustered obstacle blocks and each has two ends. By applying the rules forementioned, *B* and *F* are concave blocks.

The symbol function can be constructed as: if obstacle point belongs to a concave block, the symbol function is set to zero, as $B(i)=0$; else, $B(i)=1$. $i=0, 1, …, 180$.

### D. Construction of Threshold Function

From the view of kinematics, the main purpose of the threshold function is to set a safe distance $D_{safe}$, which makes sure that there is no crash with obstacles when the robot is moving at a certain speed.

The threshold function $H(i)$ can be obtained as follows:

$$H(i) = \begin{cases} 1, D_i > D_{safe} \\ 0, others \end{cases} \quad i=0, 1,..., 180 \quad (12)$$

### E. Construction of Cost Function

Generally, the cost functions in VPH+ methods are goal oriented and aim to find out a desired direction that makes the UGV moving closer to the goal position. Taking the heading deviation and speed into account, a time-oriented cost function for the UGV to get to the goal position in shorter time is constructed in VPH+:

$$C(i) = \frac{B(i)H(i)D_i}{S(i)} \quad (13)$$

$$S(i) = k_1 hg + k_2 ho + k_3 \quad (14)$$

Where *hg* and *ho* are the angles defined in Fig. 6; *hr* is the current heading of the robot; $k_1, k_2, k_3$ are nonzero coefficients, and $k_1 > k_2$.

The desired direction *m* for the UGV should satisfy the equation below:

$$C(m) = max(C(i)), i=0, 1, ..., 180 \quad (15)$$

### F. Design of model predictive controller

The VPH+ approach calculates an optimal moving direction when UGV makes online planning of obstacle avoidance. However, it not always kinematically available. We hereby apply a MPC controller to generate a trajetory that drive the UGV to the desired direction. Fisrt, a series of reference points are generated in the calculated direction by VPH+, then a optimal function as shown in (16) is applied to generate the optimal control action that meet the vehicle handling constraints and limits.

$$\begin{aligned} J(k) = &\sum_{i=1}^{N_p} \left\| \eta(k+i\,|\,t) - \eta_{ref}(k+i\,|\,t) \right\|_Q^2 \\ &+ \sum_{i=1}^{N_c-1} \left\| \Delta U(k+i\,|\,t) \right\|_R^2 \\ &+ \sum_{i=1}^{N_p} \left\| U(k+i\,|\,t) - U_{ref}(k+i\,|\,t) \right\|_S^2 \end{aligned} \quad (16)$$

## IV. SIMULATION EXPERIMENTS

The simulation experiments are carried out in the open source platform V-REP (Virtual Robot Experimentation Platform). All algorithm program tested are written in Matlab. The communication between the Matlab programs and the V-REP simulation platform is conducted by signal.

The UGV is build based on the vehicle model manta, as shown in Fig. 5. A LRF sensor is mounted in front of the manta vehicle. Its scanning range is of 0~180 (181 directions) and the maximum detect distance is 80 m. The UGV can partly percept the environment through the LRF sensor. We can also acquire parameters such as the position, heading, and front wheel angle of the UGV.

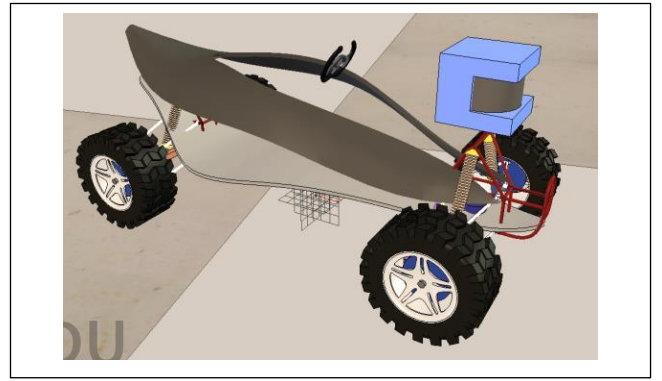

Figure 5. The UGV platform

The basic parameters of the manta model is shown in TABLE I.

TABLE I. MODEL PARAMETERS OF THE MANTA

| wheel track | 0.35 | max steer angle | 30 deg |
|---|---|---|---|
| wheel radius | 0.09 | max steering rate | 70 deg/sec |
| wheel base | 0.6 | max motor torque | 60 Nm |

A smulation scenario is designed as shown in Fig. 6. The UGV is supposed to navigate through obstacles while reaching the goal point.

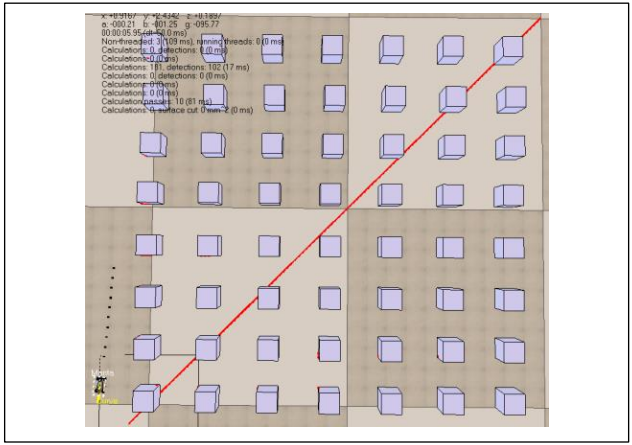

Figure 6. Simulation scenario

Both approaches can perform the assumed function, the trajectories are shown in Fig. 7.

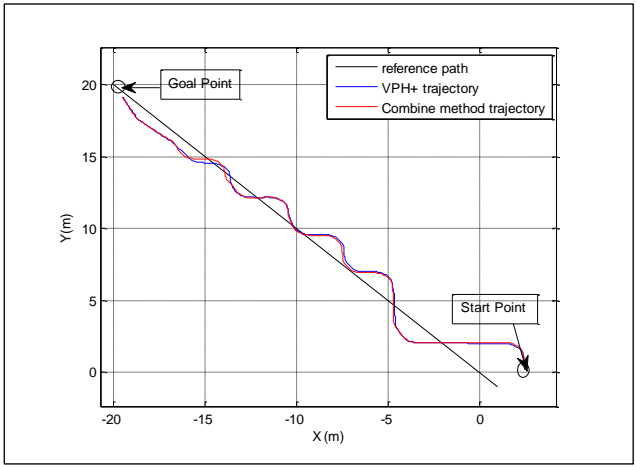

Figure 7. Tracking trajectory comparision

VPH+ approach and the proposed approach are applied respectively in this scenario. They both can perform navigation through this situation. The desired control value of the two approach is shown in Fig. 8.

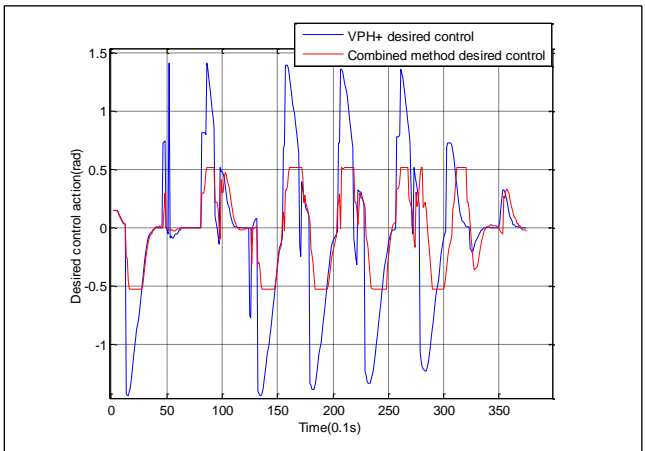

Figure 8. Desired control value comparision

The change of control action of the two approaches is shown in Fig. 9.

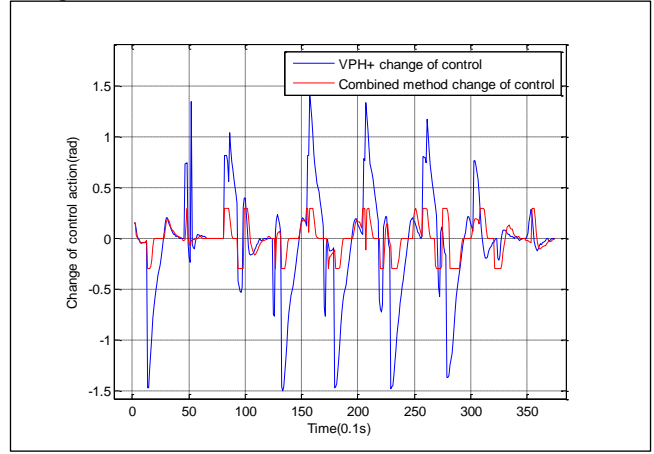

Figure 9. Change of control action comparision

As we can see from the simulation results, the proposed method can realize obstacle avoidance efficiently, meanwhile generating the control action under the limits of vhicle kinematic and dynamic constraints.

## V. CONCLUSION

This paper presented a reactive obstacle avoidance for UGV by integrating MPC with VPH+. VPH + can calculate optimal direction while MPC can perform the tracking with a model predictive controller. Simulation and experimental tests show the efectiveness of the approach. Theproposed method is tested it in different scenerrios. The sequences of the simulations demonstrated its capability of handling the scenes with obstacles.

This approach still has some drawbacks:

➢ It doesn't take the terminal constraint into consideration.

➢ We use a linearized state space model. In general we can't expect that this mathematical model produces exact predictions for the trajectories of the real real process to be controlled.

The future work will focus on the applicatio of the proposed approach on our self-driving UGV within a real environment.

## ACKNOWLEDGMENT

This research work is supported by the Natural Science Foundation of China (NSFC) (51275041). The simulations performed in VREP adopt the vehicle model design by Wangqi.